\documentclass[graybox]{svmult}

\usepackage{mathptmx}
\usepackage{helvet}
\usepackage{courier}
\usepackage{type1cm}

\usepackage{makeidx}
\usepackage{graphicx}

\usepackage{multicol}
\usepackage[bottom]{footmisc}

\usepackage{geometry}
\makeindex

\usepackage{float}
\usepackage{multirow}
\usepackage{tabularx}
\usepackage{amsmath}

\begin{document}

\title*{Adobe-MIT submission to the DSTC 4 Spoken Language Understanding pilot task}
\author{Franck Dernoncourt, Ji Young Lee, Trung H. Bui, and Hung H. Bui}
\institute{Franck Dernoncourt \at Adobe Research, San Jose, CA, USA and MIT, Cambridge, MA, USA \quad \email{francky@mit.edu}
\and Ji Young Lee \at Massachusetts Institute of Technology, Cambridge, MA, USA \quad \email{jjylee@mit.edu}
\and Trung H. Bui \at Adobe Research, San Jose, CA, USA \quad \email{bui@adobe.com}
\and Hung H. Bui \at Adobe Research, San Jose, CA, USA \quad \email{hubui@adobe.com}}
\maketitle

\vspace{-3cm}

\abstract{}
The Dialog State Tracking Challenge 4 (DSTC 4) proposes several pilot tasks. In this paper, we focus on the spoken language understanding pilot task, which consists of tagging a given utterance with speech acts and semantic slots. We compare different classifiers: the best system obtains 0.52 and 0.67 F1-scores on the test set for speech act recognition for the tourist and the guide respectively, and 0.52 F1-score for semantic tagging for both the guide and the tourist.

\vspace{-0.5cm}

\section{Speech act recognition}
\label{sec:speech-act-recognition}
\vspace{-0.2cm}
Recognizing the speech acts of the current utterance is one of the two goals of the spoken language understanding pilot task. In the training and development sets, each utterance is annotated with one speech act. One speech act is composed of zero, one or two speech act categories. Each speech act category has in turn zero, one or two speech act attributes. There are 4 speech act categories, and 22 speech act attributes. \cite{DSTC4handbook} and \cite{DSTC4} give further details on the task. The main approaches for this task are presented in \cite{ries1999hmm, ang2005automatic, stolcke2000dialogue, ji2005dialog, serafin2004flsa, zimmermann2006toward, levin1999tagging, fivsel2007machine}.

We submitted 5 systems. Systems 3 and 5 were the best performing ones. System 3 is based on a support vector machine (SVM) classifier to recognize the speech acts: the features are the 5000 most common unigrams, bigrams, trigrams, as well as a binary feature indicating whether the current speaker is different from the speaker in the last utterance. To account for the history, each feature is computed for both the current and the previous utterance. Two SVM classifiers were trained: one for each speaker. The kernel function as well as the penalty parameter of the error term were both optimized with 5-fold cross-validation. System 5 is similar, but with logistic regression as the classifier; moreover, it uses one single speaker-independent model instead of one model per speaker, as it slightly improves the results on the development set. Systems 3 and 5 assume that each utterance contains exactly one speech act category and one speech act attribute: they are therefore multiclass, monolabel classifiers, with 88 possible classes ($4 \text{ speech act categories} \times 22 \text{ speech act attributes}$).

System 4 is based on a random forest classifier and has only 4 features: the number of question marks (discrete value), whether the current speaker is different from the speaker in the last utterance, whether the current speaker is different from the speaker in the second to previous utterance, and whether the current speaker is the guide or the tourist. System 4 was designed to predict the speech act categories, but not the speech act attributes. System 2 is the same as System 4, except that System 4's features are computed on the current and previous utterances, while System 2's features are computed on the current, previous and second-to-previous utterances. 

System 1 is a rule-based classifier consisting of set of around 10 simple rules (e.g. if the preceding utterance is predicted as a question, then the current utterance is a response): it was designed to be used as a baseline. Table~\ref{tab:results} presents the results.
\vspace{-0.5cm}
\section{Semantic tagging}
\label{sec:semantic-tagging}

\vspace{-0.2cm}
Semantic tagging is the second goal of the SLU pilot task. A tagged entity comprises one or several words. A tag includes one of 8 main categories, and may contain a subcategory,  a relative modifier, and a from-to modifier. The ontology contains the list of subcategories, relative modifiers, and from-to modifiers that are present in each main category. \cite{DSTC4handbook} and \cite{DSTC4} give further details on the task. The main approaches for this task are presented in \cite{kudo2001chunking, lafferty2001conditional, raymond2007generative, mesnil2013investigation, yao2014spoken, guo2014joint, mesnil2015using, deng2012use}.

Our semantic tagging system is based on conditional random fields (CRFs) implemented by the CRFsuite library \cite{CRFsuite} and uses the following features computed on 7 consecutive words (the current word, the 3 previous words, and the 3 following words): case-insensitive unigrams, the last 3 characters of the word, whether the first letter of the word is an uppercase, whether all the letters of the word are uppercases, whether the word contains a digit, the coarse-grained part-of-speech of the word, and the fine-grained part-of-speech of the word. Four CRFs are trained independently, one for each of the 4 types of attributes: main category, subcategory, relation, and from-to. To combine the output of each CRF, a semantic tag is first generated for each sequence of words tagged by the main category CRF. The other thee attributes are included in the semantic tag if these words are tagged by the corresponding CRFs with a value that is present in the main category according to the ontology. Table~\ref{tab:results} presents the results.

\vspace{-0.3cm}
\newcolumntype{?}{!{\vrule width 1pt}}
\newcolumntype{C}{>{\centering\arraybackslash}X}%
\renewcommand*\arraystretch{1.1}
\begin{table}[h]
\caption{Results of different systems on the test set, evaluated by DSTC 4's organizers.}
\label{tab:results}
\begin{tabularx}{\textwidth}{X|CCC|CCC}
\svhline
\multirow{2}{*}{ Tracker  } & \multicolumn{3}{c|}{Guide} & \multicolumn{3}{c}{Tourist} \\
 & Precision & Recall & F1-score & Precision & Recall & F1-score  \\

\hline
System 1		& 0.6287	& 0.5191	& 0.5687	& 0.3583	& 0.2977	& 0.3252 \\
System 2		& 0.6330	& 0.5227	& 0.5726	& 0.2931	& 0.2435	& 0.2660 \\
System 3		& \textbf{0.7451}	& \textbf{0.6153}	& \textbf{0.6740}	& 0.5627	& 0.4675	& 0.5107 \\ 
System 4		& 0.6314	& 0.5214	& 0.5712	& 0.2939	& 0.2442	& 0.2668 \\	 
System 5		& 0.6762	& 0.5584	& 0.6117	& \textbf{0.5736}	& \textbf{0.4766}	& \textbf{0.5206} \\
\hline	
Semantic  		& 0.5646	& 0.4886	& 0.5239	& 0.5741	& 0.4764	& 0.5207 \\
\svhline
\end{tabularx}
\end{table}

\begin{acknowledgement}
The authors would like to warmly thank the DSTC 4 team for organizing the challenge and being so prompt to respond to emails. The authors are also grateful to the anonymous reviewers as well as to Walter Chang for their valuable feedback.
\end{acknowledgement}

\vspace{-0.9cm}
\bibliographystyle{abbrv}
\bibliography{iwsds2016}

\end{document}